\documentclass{article}
\usepackage{graphicx}
\usepackage{float}
\usepackage{times}
\usepackage{dsfont}
\usepackage{amsmath,amssymb}
\usepackage[ruled,vlined,linesnumbered]{algorithm2e}
\SetKwInput{KwInput}{Input}
\SetKwInput{KwOutput}{Output}

\SetCommentSty{texttt}
\SetKwComment{Comment}{/* }{ */}
\usepackage{multicol}
\usepackage{siunitx}
\usepackage[bookmarks=true]{hyperref}
\usepackage{cleveref}
\usepackage{xspace}
\usepackage{xurl}
\usepackage{capt-of}
\usepackage[usenames,dvipsnames]{xcolor}
\usepackage{booktabs}  
\usepackage{threeparttable}
\usepackage{multirow}
\usepackage{enumitem}
\usepackage{ulem}
\usepackage{pifont}

\usepackage{wrapfig} 
\usepackage{colortbl}
\usepackage{xcolor}
\usepackage{svg}
\usepackage{subcaption}
\usepackage[preprint]{corl_2026} 

\title{RPL: Learning Robust Humanoid Perceptive Locomotion on Challenging Terrains}

\author{
\bfseries Yuanhang Zhang$^{1,2}$, Younggyo Seo$^{1}$, Juyue Chen$^{1}$, Yifu Yuan$^{2}$, Koushil Sreenath$^{1,4}$, \\
\bfseries Pieter Abbeel$^{1,4\dagger}$, Carmelo Sferrazza$^{1\dagger}$, Karen Liu$^{1,3\dagger}$, Rocky Duan$^{1\dagger}$, Guanya Shi$^{1,2\dagger}$ \\[4pt]
\normalfont\normalsize $^{1}$Amazon FAR, $^{2}$CMU, $^{3}$Stanford University, $^{4}$UC Berkeley, $^{\dagger}$Amazon FAR Co-Lead \\
\normalfont\normalsize Page: \href{https://rpl-humanoid.github.io/}{\texttt{https://rpl-humanoid.github.io/}}
}

\definecolor{verylightgray}{RGB}{240, 240, 240}
\definecolor{llnvgreen}{RGB}{220, 237, 191}
\definecolor{rqblue}{RGB}{246, 208, 208}
\definecolor{gsblue}{RGB}{236, 178, 178}
\definecolor{gsred}{RGB}{201, 201, 247}
\definecolor{mydarkblue}{rgb}{0,0.08,0.45}
\definecolor{mydarkgreen}{RGB}{0, 139, 69}
\definecolor{mydarkpink}{RGB}{199, 21, 133}
\definecolor{mygreen2}{RGB}{0 205 0}
\definecolor{mybrown}{RGB}{139 69 19}
\definecolor{HiliteRow}{RGB}{230, 230, 255}
\definecolor{Methodred}{RGB}{191, 3, 3}
\definecolor{mydarkred}{RGB}{180,30,30}
\hypersetup{
    colorlinks=true,
    linkcolor=blue,
    urlcolor=magenta,
    citecolor=mygreen2,
}

\newcommand{\goodnumber}[1]{{\color{Methodred}\textbf{#1}}}

\newcommand{\method}{{\texttt{RPL}}\xspace}

\newcommand{\bs}[1]{\boldsymbol{#1}}

\newcommand{\pmval}[1]{{\tiny$\pm\text{#1}$}}

\begin{document}
\maketitle


\begin{figure*}[htb]
    \centering
    \vspace{-6mm}
    \includegraphics[width=\textwidth]{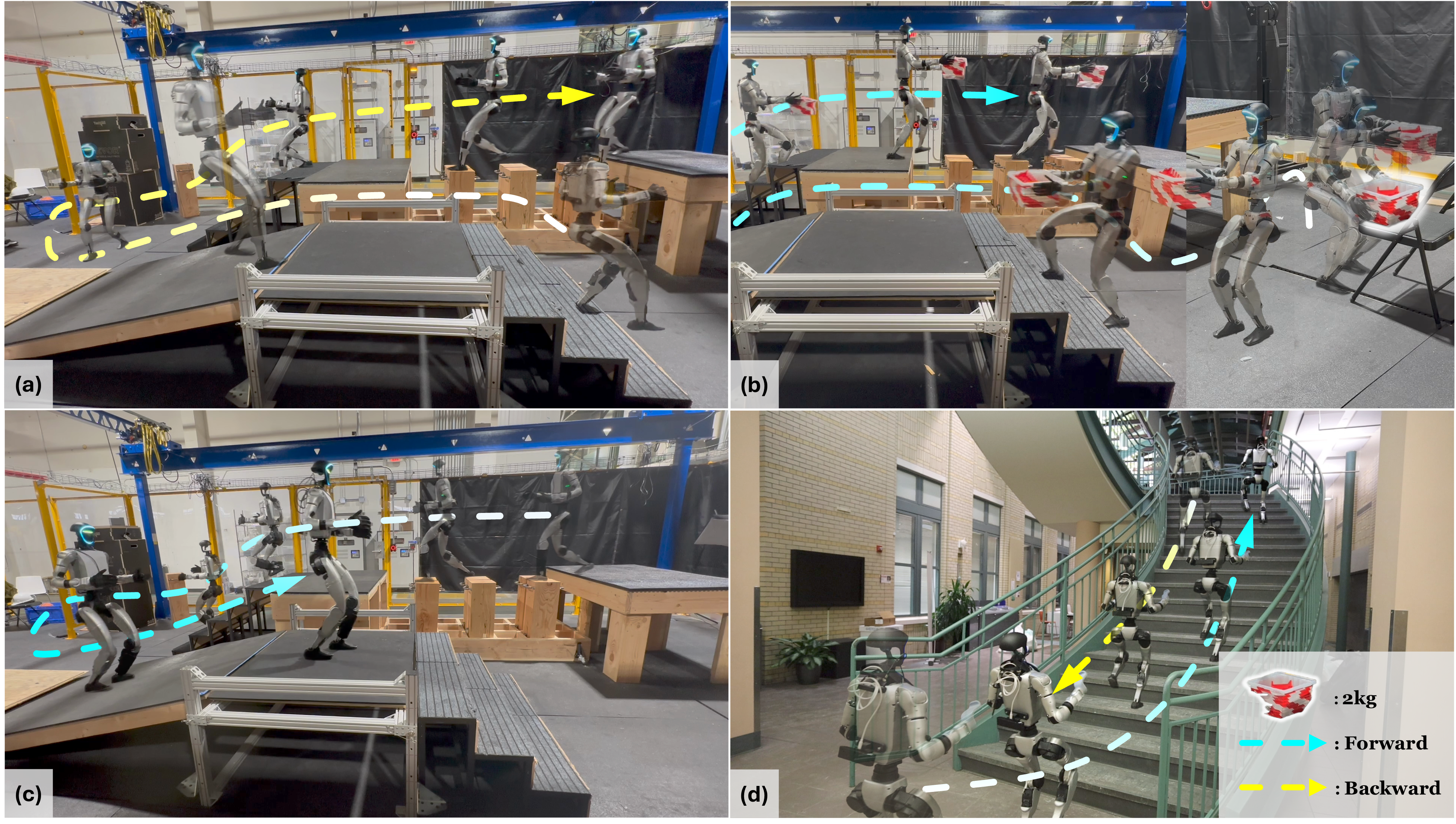}
    \vspace{-6mm}
    \caption{\small \method enables long-horizon bidirectional locomotion and remains robust with payloads on Unitree G1 humanoid robot with a depth-based transformer policy. Terrains include 20$^\circ$ slopes, staircases with different step lengths (22\,cm, 25\,cm, 30\,cm), and 25\,cm$\times$25\,cm stepping stones with 60\,cm gaps.}
    \label{fig:teaser}
\end{figure*}
\vspace{-2mm}
\begin{abstract}
Humanoid perceptive locomotion has made significant progress and shows great promise, yet achieving robust multi-directional locomotion on complex terrains remains underexplored. To tackle this challenge, we propose \method, a two-stage training framework that enables multi-directional locomotion on challenging terrains, and remains robust with payloads. \method first trains terrain-specific expert policies with privileged height map observations to master decoupled locomotion and manipulation skills across different terrains, and then distills them into a transformer policy that leverages multiple depth cameras to cover a wide range of views. During distillation, we introduce two techniques to robustify multi-directional locomotion, depth feature scaling based on velocity commands and random side masking, which are critical for asymmetric depth observations and unseen widths of terrains. For scalable depth distillation, we develop an efficient multi-depth system that ray-casts against both dynamic robot meshes and static terrain meshes in massively parallel environments, achieving a 5$\times$ speedup over the depth rendering pipelines in existing simulators while modeling realistic sensor latency, noise, and dropout. Extensive real-world experiments demonstrate robust multi-directional locomotion with payloads (2\,kg) across challenging terrains, including 20$^\circ$ slopes, staircases with different step lengths (22\,cm, 25\,cm, 30\,cm), and 25\,cm$\times$25\,cm stepping stones separated by 60\,cm gaps.
\end{abstract}
\keywords{Humanoid Perceptive Locomotion, RL, Imitation Learning}

\section{Introduction}
\label{sec:introduction}

Humanoid robots hold great potential for daily human tasks. While recent work has demonstrated expressive whole-body motion tracking and loco-manipulation capabilities~\cite{zhang2025falcon,homie,ji2024exbody2,cheng2024expressive,sombolestan2024adaptive,fey2025bridging,he2024omnih2o,he2025asap,li2023dynamic,murooka2021humanoid,bouyarmane2018quadratic,fu2024humanplus,yang2025omniretarget}, a defining capability beyond wheeled mobile manipulators is leveraging legged mobility to robustly traverse diverse terrains, while performing meaningful tasks such as carrying payloads.
Existing humanoid perceptive locomotion methods~\cite{wang2025beamdojo,cheng2024extreme,zhuang2024humanoid,long2025learning,ben2025gallant,hoeller2024anymal,sun2025dpl,sun2025learning,wang2025omni,yang2021learning,he2025attention} typically rely on a single forward camera and idealized upper-body conditions, becoming brittle under multi-directional locomotion or dynamic self-occlusions from upper-body motions. Mapping-based methods~\cite{miki2022learning,wang2025omni,long2025learning,ben2025gallant,wang2025beamdojo} can handle bidirectional locomotion on challenging terrains but rely on noisy state estimation that complicates system-level optimization. End-to-end perceptive control under asymmetric multi-view inputs (i.e., different cameras observing different terrain types) remains underexplored.

We propose \method, a two-stage framework for \textbf{R}obust humanoid \textbf{P}erceptive \textbf{L}ocomotion. \method first trains terrain-specific expert policies with privileged height maps under end-effector force perturbations, acquiring decoupled locomotion and manipulation skills. These experts are then distilled into a unified transformer policy taking multi-view depth, enabling robust multi-directional locomotion. For scalable distillation, we develop an efficient multi-depth rendering system ray-casting against both dynamic robot meshes and static terrain meshes in massively parallel environments, modeling realistic latency, noise, and dropout. We further introduce two distillation techniques: (1) \textit{Depth Feature Scaling based on Velocity commands} (DFSV), which adaptively modulates perception features based on the velocity command to reduce distribution shift under asymmetric multi-view inputs; and (2) \textit{Random Side Masking} (RSM), which randomly masks lateral regions of depth images to improve generalization to unseen terrain widths.

Our key contributions are: (1) \method, a two-stage framework distilling terrain-specific experts into a unified depth policy for robust multi-directional humanoid locomotion; (2) an efficient multi-depth rendering system over dynamic robot and static terrain meshes, achieving a 5$\times$ speedup over existing simulators with realistic latency, noise, and dropout; (3) Depth Feature Scaling based on Velocity commands (DFSV) and Random Side Masking (RSM) during distillation to handle asymmetric multi-view inputs and unseen terrain widths; and (4) real-world validation of long-horizon bidirectional locomotion under whole-body motions and payload disturbances.

\section{Related Works}
\label{sec:relatedwork}

\subsection{Perceptive Locomotion in Legged Robots}
Perceptive locomotion for legged robots has progressed rapidly~\cite{miki2022learning,wang2025beamdojo,cheng2024extreme,zhuang2024humanoid,long2025learning,hoeller2024anymal,sun2025dpl,sun2025learning,wang2025omni,yang2021learning,he2025attention,ben2025gallant,zhu2026hiking,agarwal2023legged}. Quadrupeds can sometimes traverse stairs and curbs blindly~\cite{lee2020learning} thanks to inherently stable support polygons, whereas humanoids demand precise foot placement and are far more sensitive to perception errors. Two paradigms dominate: (1) LiDAR-based mapping~\cite{miki2022learning,wang2025beamdojo,long2025learning,ben2025gallant,hoeller2024anymal,he2025attention,wang2025omni} compresses terrain into elevation maps but relies on calibration and state estimation that introduce sim-to-real drift; (2) end-to-end depth-based methods~\cite{cheng2024extreme,sun2025dpl,sun2025learning,agarwal2023legged,zhuang2024humanoid,zhu2026hiking,yang2021learning} map raw depth directly to actions, removing explicit estimation.

Most depth-based work uses a single forward camera and demonstrates only forward locomotion. \cite{song2025gait} achieves omnidirectional traversal with a single downward camera but struggles with sparse footholds (e.g., stepping stones); \cite{gadde2025no} adds multiple cameras for wider coverage but evaluates only on stairs, not the general asymmetric multi-view setting where different cameras observe distinct terrain types. Depth rendering also remains expensive: recent Warp-based pipelines~\cite{zhu2026hiking,wang2025omni,sun2025dpl,wang2025more} accelerate it but lack dynamic-mesh support or do not scale to multiple cameras. Existing policies also rarely evaluate robustness to unseen terrain widths or loco-manipulation disturbances such as transporting payloads. \method targets this gap, combining scalable multi-depth rendering with distillation techniques that generalize across unseen terrain widths under payloads.

\begin{table}[!ht]
\centering
\begingroup
\footnotesize
\setlength{\tabcolsep}{3pt}
\renewcommand{\arraystretch}{1.0}
\newcommand{\pwyes}{\textcolor{mydarkgreen}{\checkmark}}
\newcommand{\pwno}{\textcolor{red}{$\times$}}
\newcommand{\pwpartial}{\textcolor{orange}{$\sim$}}
\newcommand{\pwna}{\textcolor{gray}{N/A}}
\resizebox{\textwidth}{!}{%
\begin{tabular}{lcccccc}
\toprule
\textbf{Method} & \shortstack{\textbf{Stairs}\\\textbf{Up/Down}} & \shortstack{\textbf{Stepping}\\\textbf{stones}} & \shortstack{\textbf{Multi-view}\\\textbf{coverage}} & \shortstack{\textbf{Real-world}\\\textbf{Bidirectional}} & \shortstack{\textbf{Manip. /}\\\textbf{Active Payload}} & \shortstack{\textbf{Sim Traversal}\\slopes/stairs/stones} \\
\midrule
\rowcolor{yellow!12}
\textbf{RPL (ours)} & \pwyes & \pwyes & \pwyes & \pwyes & \pwyes & \pwyes/\pwyes/\pwyes \\
Hiking-in-the-Wild~\cite{zhu2026hiking} & \pwyes & \pwno & \pwno & \pwno & \pwyes & \pwyes/\pwpartial/\pwno \\
DPL~\cite{sun2025dpl} & \pwyes & \pwno & \pwno & \pwno & \pwyes & \pwna \\
Gait-adaptive~\cite{song2025gait} & \pwyes & \pwno & \pwyes & \pwyes & \pwyes & \pwna \\
No More Blind Spots~\cite{gadde2025no} & \pwyes & \pwno & \pwyes & \pwyes & \pwno & \pwna \\
BeamDojo~\cite{wang2025beamdojo} & \pwyes & \pwyes & \pwyes & \pwyes & \pwno & \pwna \\
Humanoid Parkour~\cite{zhuang2024humanoid} & \pwyes & \pwno & \pwno & \pwno & \pwyes & \pwna \\
\bottomrule
\end{tabular}}
\caption{\footnotesize \textsc{Capability comparison vs. related methods.} \pwyes\ yes, \pwpartial\ partial (only single-direction), \pwno\ no. \textit{Multi-View}: 360 deg LiDAR or multi-camera; \textit{Sim Traversal}: single-shot per-terrain success (slopes/stairs/stones); N/A marks without public code.}
\label{tab:prior_work}
\endgroup
\vspace{-4mm}
\end{table}

\textbf{Comparison with prior work.} \Cref{tab:prior_work} highlights \method's combination of onboard multi-view depth, real-world bidirectional traversal, and teleoperated active payload transport across stairs, slopes, and stepping stones.

\begin{figure*}[tb]
    \centering
    \includegraphics[width=\textwidth]{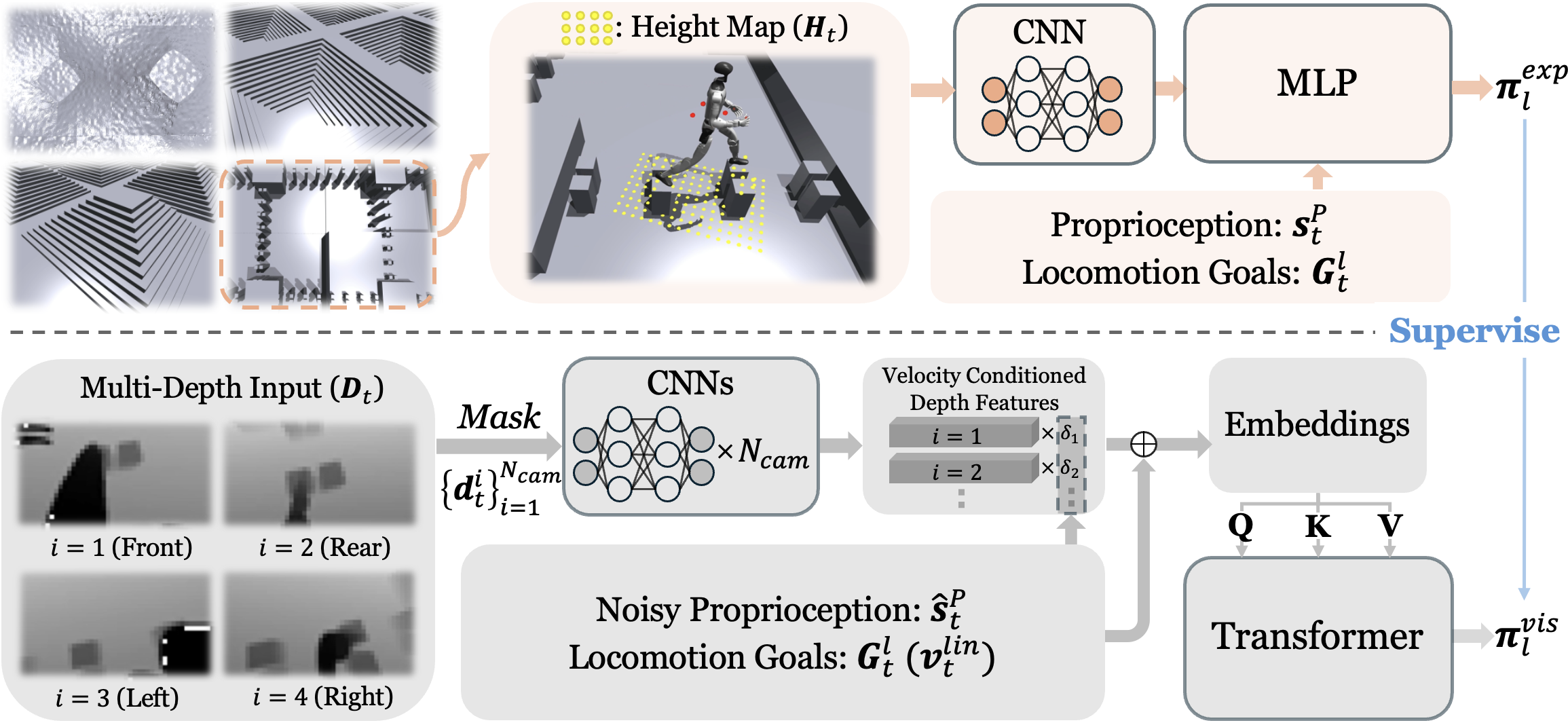}
    \vspace{-4mm}
    \caption{\small
    \textbf{Overview of the Two-Stage Training Framework in \method}.
    \textbf{Top (Stage 1)}: We train terrain-specific expert policies using privileged height-map observations to master decoupled locomotion and manipulation.
    \textbf{Bottom (Stage 2)}: We distill the expert policies into a single multi-view, depth-based transformer policy. Multi-camera depth inputs are encoded by CNN backbones with random side masking (\Cref{sec:rsm}) and depth feature scaling based on velocity commands (\Cref{sec:dfsv}), fused into visual features, and combined with noisy proprioceptive observations and task goals to predict student actions.
    }
    \label{fig:method}
    \vspace{-5mm}
\end{figure*}

\section{\method: Two Stage -- From Multiple Experts to One General Visual Policy}
\label{sec:falcon}

\method is a two-stage framework (\Cref{fig:method}). Stage~1 trains terrain-specialized experts with privileged height-map observations over slopes, stairs up, stairs down, and stepping stones (\Cref{fig:terrains}). Stage~2 distills them into a single unified visual policy taking front and back depth observations for bidirectional locomotion. We build on the dual-agent formulation of FALCON~\cite{zhang2025falcon}, which factorizes whole-body control into a lower-body locomotion policy and an upper-body manipulation policy on a shared proprioceptive history, outperforming single-agent whole-body control~\cite{zhang2025falcon,li2025hold,ding2025jaeger}.
Both agents observe the proprioceptive history $\mathbf{s}_t^p = [\mathbf{q}_{t-4:t}, \dot{\mathbf{q}}_{t-4:t}, \boldsymbol{\omega}^{\text{root}}_{t-4:t}, \mathbf{g}_{t-4:t}, \mathbf{a}_{t-5:t-1}]$. The lower-body policy is conditioned on locomotion goals $\mathbf{G}_t^l = [\mathbf{v}_t^{\text{lin}}, \mathbf{w}_t^{\text{yaw}}, \boldsymbol{\phi}_t^{\text{stance}}, \mathbf{h}_t^{\text{root}}, \mathbf{o}_t^{\text{torso}}]$ (linear/angular velocity, stance/walking mode, root height, torso orientation; root height and torso are tracked only during stance), and the upper-body policy on $\mathbf{G}_t^u = \mathbf{q}_{\text{upper},t}$ (upper-body joint targets).
The perceptual observation is $\mathbf{H}_t$ in Stage~1 (privileged height map) and $\mathbf{D}_t=\{\mathbf{d}_t^{(i)}\}_{i=1}^{N_{\text{cam}}}$ in Stage~2 (multi-view depth), fed only to the lower-body policy since terrain perception primarily governs legged locomotion. The policies are $\boldsymbol{\pi}_l:(\mathbf{s}_t^p, \mathbf{G}_t^l, \mathbf{H}_t \,\mathrm{or}\, \mathbf{D}_t) \!\rightarrow\! \mathbf{a}_t^l$ and $\boldsymbol{\pi}_u:(\mathbf{s}_t^p, \mathbf{G}_t^u) \!\rightarrow\! \mathbf{a}_t^u$, with concatenated actions $\mathbf{a}_t = [\mathbf{a}_t^l;\mathbf{a}_t^u]$ tracked by a low-level PD controller. (See Appendix~\ref{appendix:notation} for the notation summary.)

\subsection{Stage 1: Training Multiple Terrain Experts via Height Map}
\label{subsec:expert}

To provide high-quality teacher actions for distillation, we train one terrain-specialized expert per terrain family using a privileged height map represented as a local grid of size $1.6\,\mathrm{m}\times1.0\,\mathrm{m}$ at $0.1\,\mathrm{m}$ resolution. The dual-agent policies are jointly trained with independent reward functions and optimized using PPO~\cite{PPO}. We adopt a force curriculum~\cite{zhang2025falcon} with direction ratio $\boldsymbol{\gamma}=[\gamma_x,\gamma_y,\gamma_z]=[0,0,1]$ to specifically address end-effector payloads for robust loco-manipulation. We further apply asymmetric actor--critic training~\cite{pinto2017asymmetric}, where the critics access privileged information including root linear velocity and end-effector force $\bs{F}^{ee}_t$, and use symmetry data augmentation~\cite{mittal2024symmetry} on $\mathbf{s}_t^p$, $\mathbf{G}_t^l$, $\mathbf{G}_t^u$, and $\mathbf{H}_t^l$ to encourage symmetric gait patterns.

\subsubsection{Terrain Settings}
We train specialized experts for four terrain families illustrated in~\Cref{fig:method}:
(1) \textbf{Slopes}: ramps with inclinations up to $37^\circ$ in pyramid form;
(2) \textbf{Stairs Up~/~Down}: ascending and descending staircases with $0.25$--$0.30\,\mathrm{m}$ step length and $0.05$--$0.27\,\mathrm{m}$ height;
(3) \textbf{Stepping Stones}: column-shaped discrete footholds with $0.25$--$0.40\,\mathrm{m}$ diameter, $0.05$--$0.70\,\mathrm{m}$ gaps, and up to $0.05\,\mathrm{m}$ height variation. Stairs and stepping stones are particularly challenging given the robot's $0.21\,\mathrm{m}$ foot length, which is close to the minimum support size of $0.25\,\mathrm{m}$, leaving little margin for foothold placement errors.
\subsubsection{Reward Design} 
\begin{figure}[tb]
    \centering
    \begin{minipage}[b]{0.49\columnwidth}
        \centering
        \includegraphics[width=\linewidth]{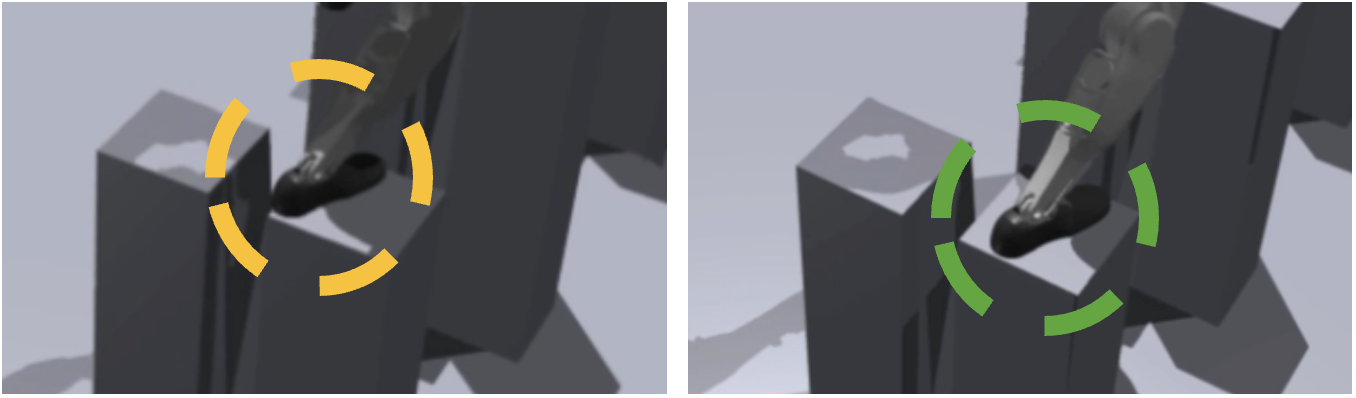}
        \caption{\small w/o (left) and w/ (right) foot-edge penalty.}
        \label{fig:footedge}
    \end{minipage}\hfill
    \begin{minipage}[b]{0.49\columnwidth}
        \centering
        \includegraphics[width=\linewidth]{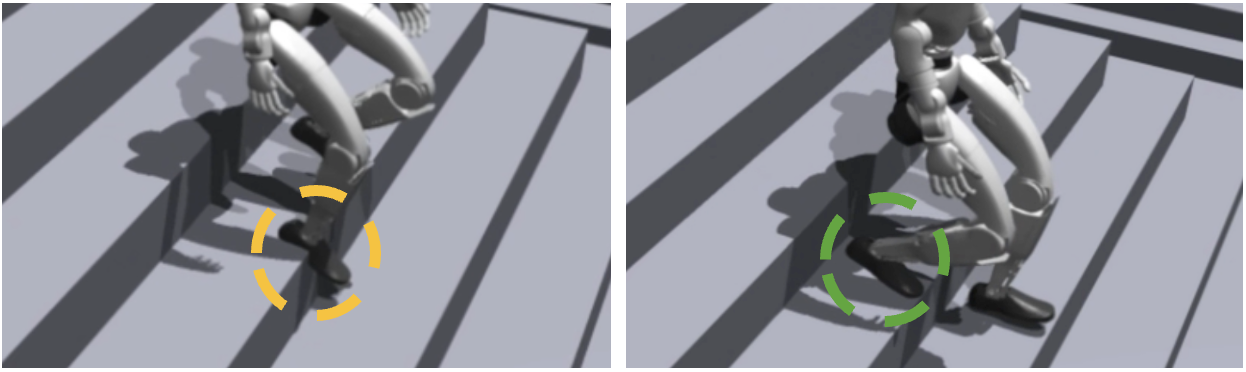}
        \caption{\small w/o (left) and w/ (right) foothold penalty.}
        \label{fig:foothold}
    \end{minipage}
    \vspace{-4mm}
\end{figure}

We adopt the dual-agent rewards in~\cite{zhang2025falcon} and add three terms for stable terrain traversal: (i) \textbf{Foot Edge Penalty}~\cite{cheng2024extreme}, which uses precomputed dilated edge masks to penalize contacts at terrain edges (\Cref{fig:footedge}); (ii) \textbf{Foothold Penalty}~\cite{wang2025beamdojo}, which samples a grid across the foot sole and penalizes invalid coverage (\Cref{fig:foothold}); and (iii) \textbf{Torso Orientation Tracking}~\cite{li2025amo}, $r_{torso} = \exp(-\|\mathbf{g}_{\text{proj}} - \mathbf{g}_{\text{ref}}\|^2 / \sigma)$, coordinating waist and hip joints for a wide stance-mode tracking range. We use foothold penalty for stairs (edge dilation is too conservative on 0.25\,m treads vs. 0.21\,m feet) and foot-edge penalty for stepping stones (small isolated supports offer little tolerance to placement errors).

\subsection{Stage 2: Distillation into Unified Visuomotor Policy}
\label{subsec:student}

As shown in~\Cref{fig:method}, after all the terrain experts based on height maps are obtained, we distill them into one unified visuomotor policy for multi-directional perceptive locomotion.

\subsubsection{Distillation Formulation}

We use DAgger~\cite{ross2011dagger} style imitation learning to train the student policy. The perceptual observation $\mathbf{P}_t^l$ consists of depth inputs from multiple depth cameras. The student lower-body policy $\boldsymbol{\pi}_l^{\text{vis}}$ takes noisy proprioception $\hat{\mathbf{s}}_t^p$ and multi-view depth inputs $\mathbf{D}_t=\{\mathbf{d}_t^{(i)}\}_{i=1}^{N_{\text{cam}}}$, while the expert policies $\boldsymbol{\pi}_{l,k}^{\text{exp}}$ are conditioned on noise-free proprioception $\mathbf{s}_t^p$ and privileged height maps $\mathbf{H}_t$.
We train the student by minimizing the action regression loss:

\begin{equation}
\mathcal{L}_{\text{distill}}
=
\mathbb{E}_{k,t}\!\left[
\left\|
\boldsymbol{\pi}_l^{\text{vis}}(\mathbf{s}_t^p,\mathbf{G}_t^l,\mathbf{D}_t)
-
\boldsymbol{\pi}_{l,k}^{\text{exp}}(\mathbf{s}_t^p,\mathbf{G}_t^l,\mathbf{H}_t)
\right\|_2^2
\right].
\label{eq:distill_loss}
\end{equation}

The final deployed controller combines the distilled visual locomotion policy $\boldsymbol{\pi}_l^{\text{vis}}$ with the blind upper-body policy $\boldsymbol{\pi}_u$, producing the whole-body action for low-level PD tracking. We use DAgger Only rather than a Hybrid DAgger\,+\,RL objective, which we find preserves cross-terrain gradient alignment and achieves higher terrain levels on every terrain family (See Appendix \ref{appendix:dagger_vs_hybrid} for details).

\subsubsection{Efficient Multi-Depth System Simulation}
\label{sec:depth_rendering}

We implement a GPU-efficient multi-depth camera simulation in NVIDIA Warp~\cite{macklin2022warp}, performing massively parallel depth synthesis by ray-casting against both dynamic robot meshes and a shared static terrain mesh (\Cref{alg:depth_raycast}). For each environment $e$, camera $c$, and pixel $(x,y)$ with intrinsics $\mathbf{K}_e$, we form a camera-frame ray $\mathbf{r}_c = \mathbf{K}_e^{-1}[x, y, 1]^\top$, rotate it by the camera orientation $\mathbf{q}_c^{e,c}$, and cast from the origin $\mathbf{o}=\mathbf{p}_c^{e,c}$ along direction $\mathbf{d}$. The depth is the distance to the closest ray--mesh intersection, clipped by far-plane $d_{\max}$.

To avoid costly per-frame mesh refitting~\cite{kulkarni2025aerial}, we keep each body mesh $\mathcal{M}_b$ canonical in its local frame and transform rays as $\mathbf{o}_b = \mathbf{R}_b^{\top}(\mathbf{o} - \mathbf{t}_b)$, $\mathbf{d}_b = \mathbf{R}_b^{\top}\mathbf{d}$, where $(\mathbf{t}_b,\mathbf{R}_b)$ is the body pose. For each ray, we first query all robot body meshes locally—tracking the closest hit $z^\star$ as an early-termination upper bound—and then the shared terrain mesh in the world frame, updating $z^\star$ if a closer hit is found. All queries execute in a single fused Warp kernel parallelized over environments, cameras, and pixels, and the entire pipeline is captured into a CUDA graph for low-overhead replay; per-environment intrinsics randomization is supported by batching $\{\mathbf{K}_e^{-1}\}$. The full kernel is summarized in~\Cref{alg:depth_raycast}.

\begin{algorithm}[H]
\caption{Multi-Depth Ray Casting Kernel}
\label{alg:depth_raycast}

\KwInput{Static terrain mesh $\mathcal{M}_{\text{terrain}}$; robot body meshes $\{\mathcal{M}_b\}_{b=1}^{B}$; camera poses $\{(\mathbf{p}_c^{e,c}, \mathbf{q}_c^{e,c})\}$; body poses $\{(\mathbf{t}_b^{e}, \mathbf{q}_b^{e})\}$; batched inverse intrinsics $\{\mathbf{K}^{-1}_e\}$; far plane $d_{\max}$; principal pixel $(c_x,c_y)$ }
\KwOutput{Depth image $\mathbf{D}\in\mathbb{R}^{N\times C\times H\times W}$}

\ForEach{$(e,c,x,y)\in [N]\times[C]\times[W]\times[H]$}{
    $(\mathbf{o},\mathbf{d},m)\gets \textsc{BuildDepthRay}(\mathbf{p}_c^{e,c},\mathbf{q}_c^{e,c},\mathbf{K}_e^{-1},x,y,c_x,c_y)$\;
    $z^\star \gets d_{\max}$; $d_{\text{bound}} \gets d_{\max}$\;

    \Comment{1) Query dynamic robot bodies}
    \For{$b\gets 1$ \KwTo $B$}{
        $(\mathbf{o}_l,\mathbf{d}_l)\gets \textsc{WorldToBodyRay}(\mathbf{o},\mathbf{d},\mathbf{t}_b^{e},\mathbf{q}_b^{e})$\;
        \If{$\textsc{MeshRayQuery}(\mathcal{M}_b,\mathbf{o}_l,\mathbf{d}_l,\,d_{\text{bound}}/m)\rightarrow t$}{
            $z^\star \gets \min(z^\star,\, m\,t)$\;
            $d_{\text{bound}} \gets z^\star$ \tcp*[r]{early termination}
        }
    }

    \Comment{2) Query static terrain}
    \If{$\textsc{MeshRayQuery}(\mathcal{M}_{\text{terrain}},\mathbf{o},\mathbf{d},\,d_{\text{bound}}/m)\rightarrow t$}{
        $z^\star \gets m\,t$\;
    }

    $\mathbf{D}[e,c,y,x] \gets z^\star$\;
}
\end{algorithm}

\subsubsection{Depth Feature Scaling Based on Velocity Commands}
\label{sec:dfsv}
\begin{figure}[t]
    \centering
    \begin{subfigure}[b]{0.53\columnwidth}
        \centering
        \includegraphics[width=\linewidth]{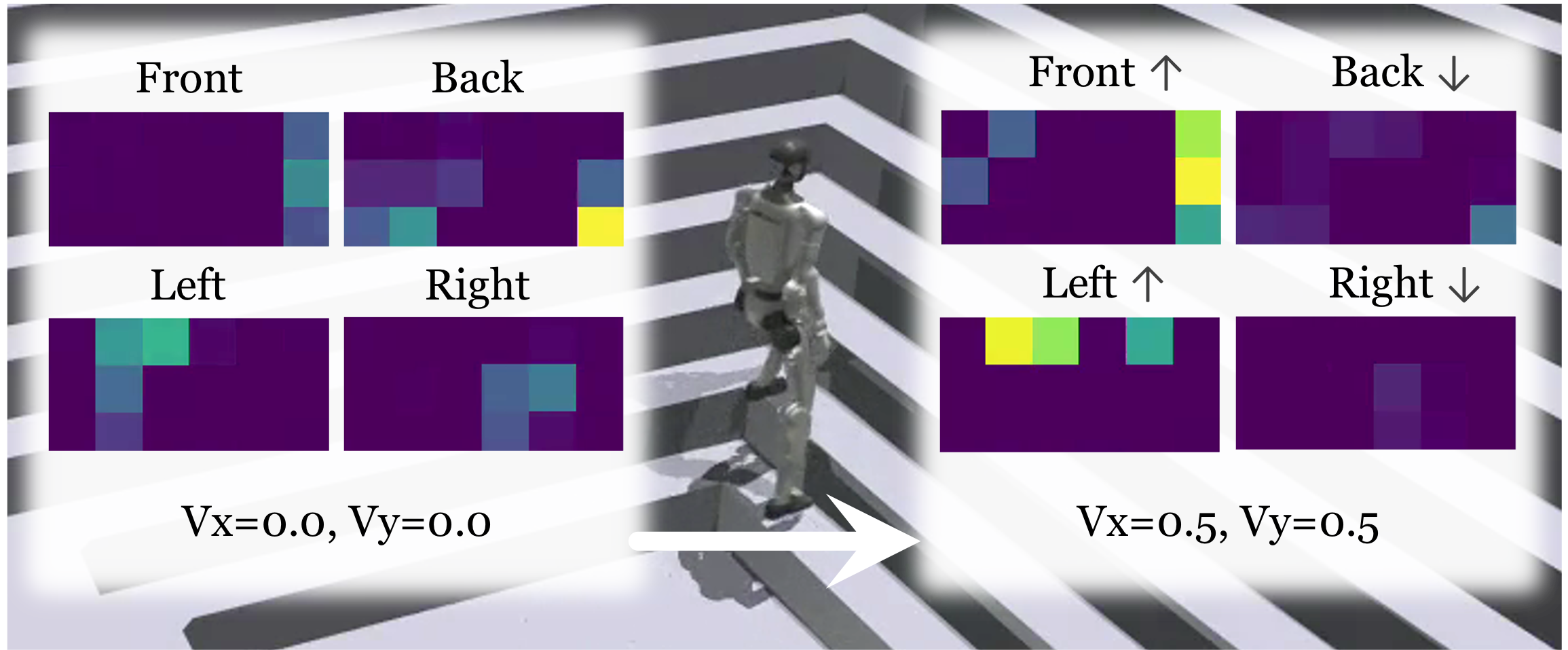}
        \caption{\small Illustration of DFSV.}
        \label{fig:dfsv}
    \end{subfigure}
    \begin{subfigure}[b]{0.44\columnwidth}
        \centering
        \includegraphics[width=\linewidth]{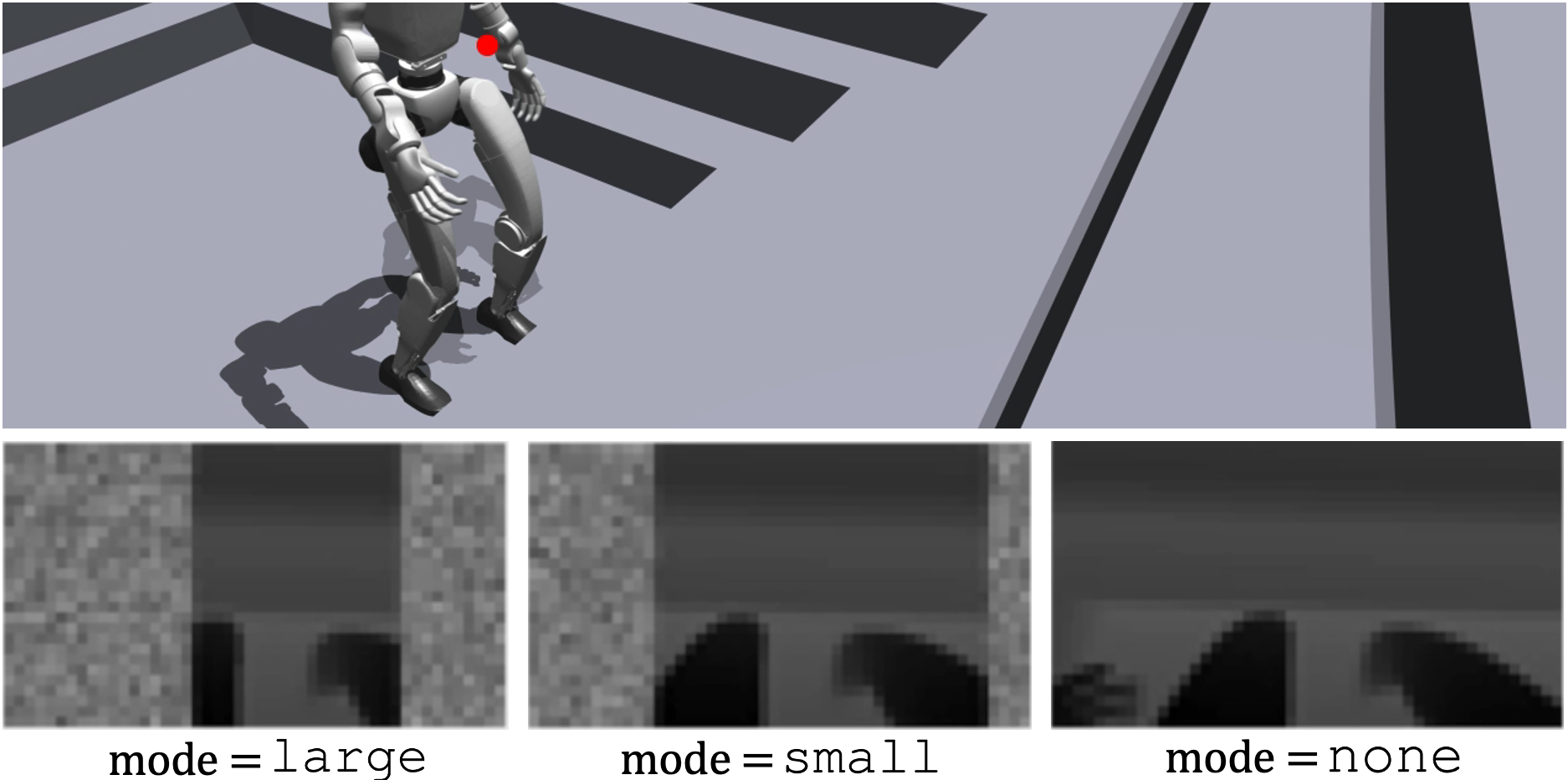}
        \caption{\small Illustration of RSM.}
        \label{fig:rsm}
    \end{subfigure}
    \vspace{-2mm}
    \caption{\small (a) \textbf{DFSV.} CNN feature maps from four directional depth cameras under a zero command ($v_x{=}0, v_y{=}0$, left) and a diagonal command ($v_x{=}0.5, v_y{=}0.5$, right); heatmaps visualize each camera's relevance to the commanded velocity direction. (b) \textbf{RSM.} The red dot on the top figure is where the front camera is placed; the bottom three figures show different masking modes.}
    \label{fig:dfsv_rsm}
    \vspace{-2mm}
\end{figure}

When the robot is equipped with multiple depth cameras facing different directions, the relevance of each camera depends on the locomotion direction. We thus introduce Depth Feature Scaling based on Velocity commands (DFSV, \Cref{fig:dfsv}). Let $\boldsymbol{\hat{n}}_i \in \mathbb{R}^2$ denote the normalized viewing direction of camera $i$ in the robot's local frame. Given the planar velocity command $\mathbf{v}^{lin}_l = [v_x, v_y]^\top$, we compute per-camera attention scales $\delta_i = 1 - \sigma(-k(\langle \mathbf{v}^{lin}_{l}, \boldsymbol{\hat{n}}_i \rangle - v_{\text{th}}))$ for $i = 1, \ldots, N_{\text{cam}}$, where $k$ controls the transition sharpness and $v_{\text{th}}$ is a velocity threshold. Cameras aligned with the commanded velocity are emphasized ($\delta_i \approx 1$), while those facing away are suppressed ($\delta_i \approx 0$). The scales are applied element-wise to the CNN feature vectors $\mathbf{f}_i$ before concatenation: $\mathbf{f}_{\text{fused}} = \bigoplus_{i=1}^{N_{\text{cam}}} \delta_i \cdot \mathbf{f}_i$, requiring no additional learned parameters.

\subsubsection{Random Side Masking for Unseen Terrain Widths}
\label{sec:rsm}
Real-world terrains often have widths different from the wide terrains seen during training. To generalize to narrower ones, we introduce Random Side Masking (RSM, \Cref{fig:rsm}): during training we stochastically occlude peripheral regions of the depth image with random noise, forcing the policy to rely on the central visible region. For each environment, we sample $\mathrm{mode} \sim \mathrm{Categorical}(\mathbf{p}_k)$ over $\{\texttt{none}, \texttt{small}, \texttt{large}\}$, where $\mathbf{p}_k$ depends on the terrain type $k$; larger masks emulate narrower effective widths. Occluded pixels are filled with uniformly sampled values in $[d_{\text{near}}, d_{\text{far}}]$ to prevent exploitation of boundary artifacts. We use larger masks on continuous terrains (stairs, slopes), where traversability is governed by local geometry ahead, and smaller masks on stepping stones, where valid footholds may appear only peripherally.

\section{Experiments}
\label{sec:experiments}
In this section, we will present extensive experiments in both simulation and the real world to show the effectiveness and the robustness of our method \method. We aim to answer four questions: 
(Q1) Does \method enable scalable multi-depth rendering compared to existing simulation methods? 
(Q2) Do DFSV and RSM improve robustness under asymmetric visual inputs and unseen terrain widths? 
(Q3) Does our network architecture outperform alternatives in distillation? 
(Q4) Can \method robustly perform long-horizon bidirectional locomotion across diverse, challenging terrains in the real world?

\subsection{Experiment Setup}
\textbf{Training Configurations.} We use Unitree G1 29 DoF as the humanoid and train all policies in NVIDIA IsaacGym~\cite{isaacgym}. Stage~1 uses 4 NVIDIA L40S GPUs with $4096\times4$ parallel envs for 12h for each expert policy; Stage~2 uses 8 GPUs with $1024\times8$ envs for 12h.

\begin{figure}[tb]
    \centering
    \begin{subfigure}[b]{0.45\columnwidth}
        \centering
        \includegraphics[width=\linewidth]{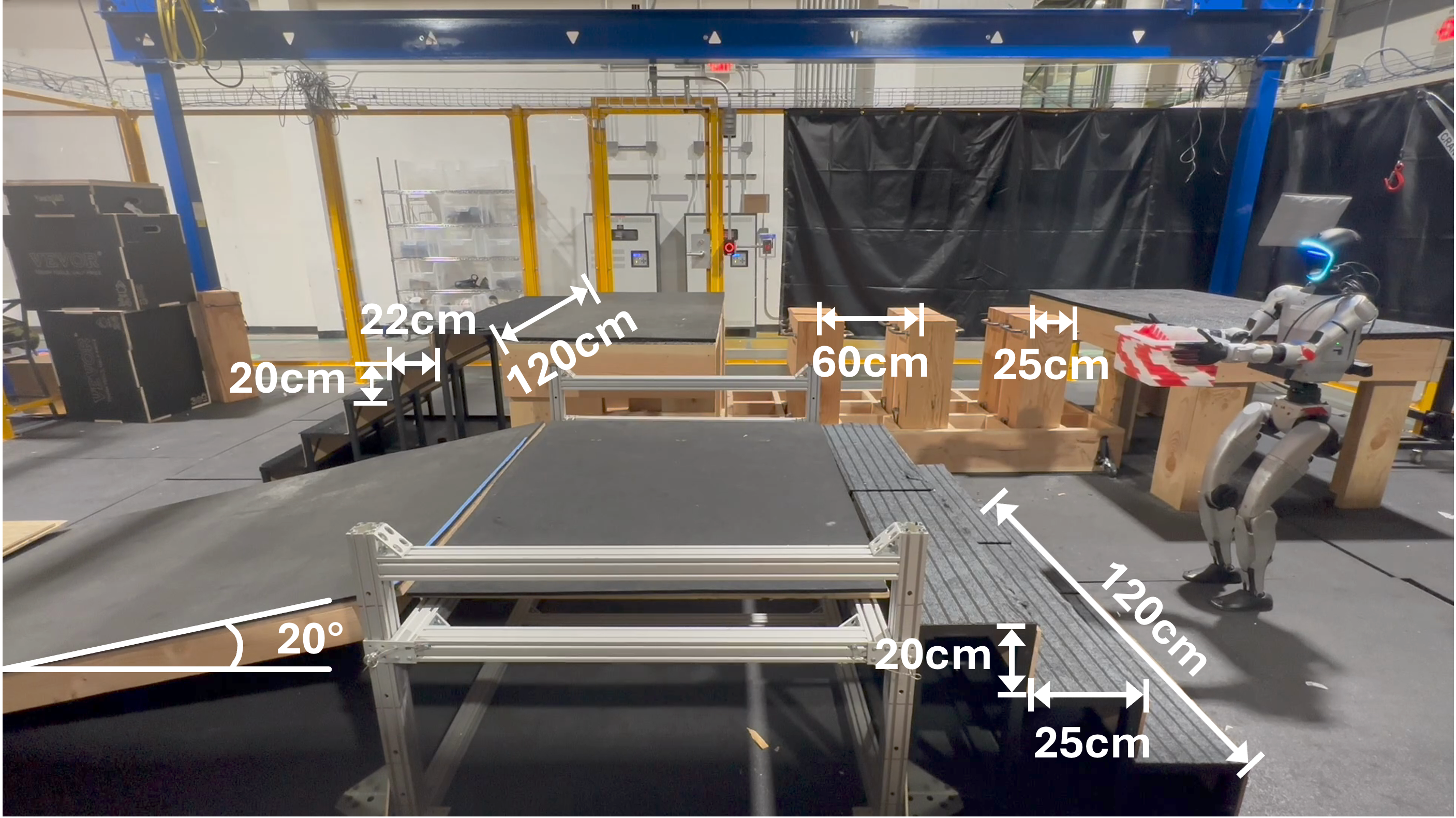}
        \caption{\small Real-world terrain course overview.}
        \label{fig:terrains}
    \end{subfigure}
    \begin{subfigure}[b]{0.52\columnwidth}
        \centering
        \includegraphics[width=\linewidth]{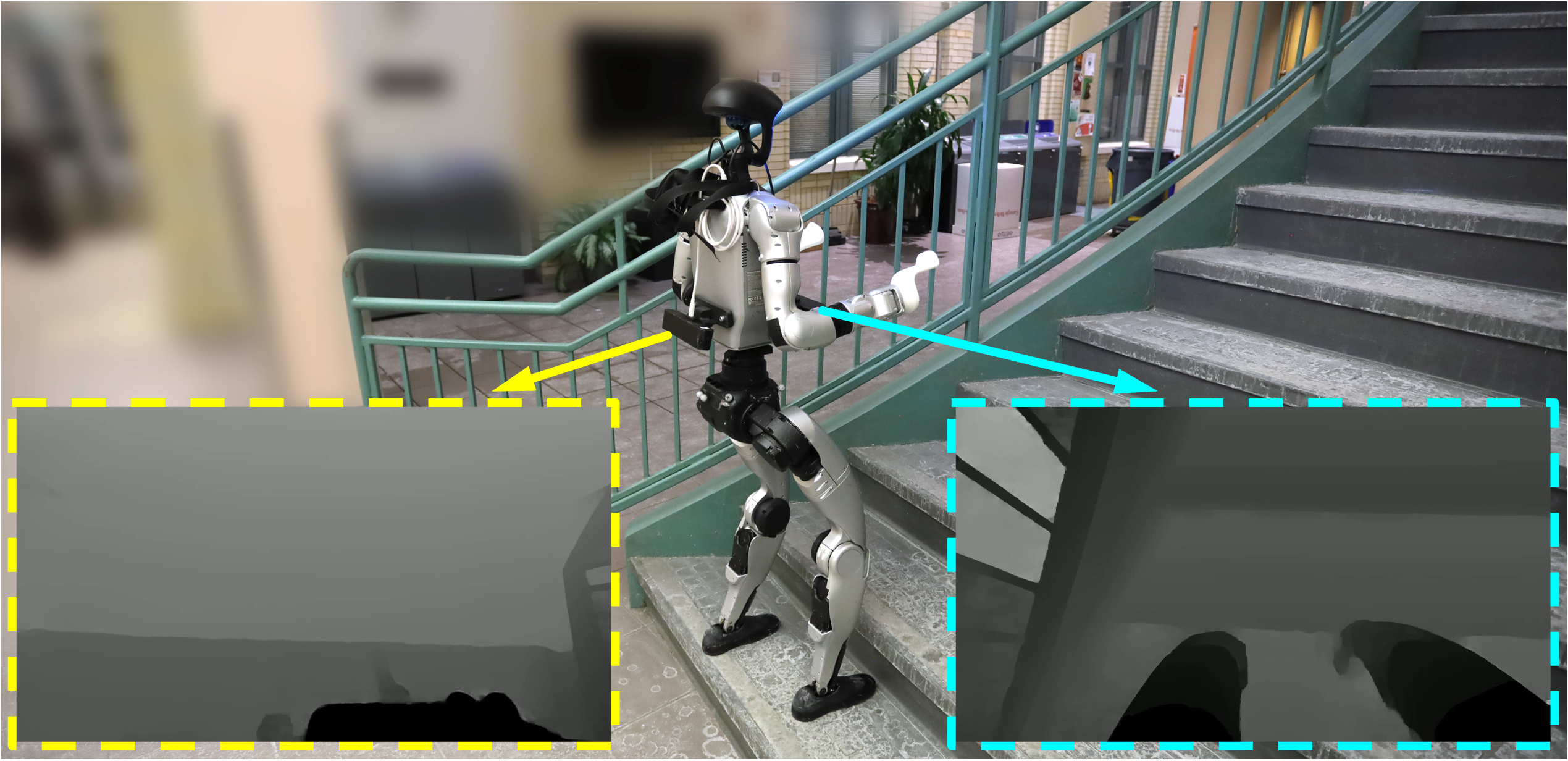}
        \caption{\small Robot setup.}
        \label{fig:robot_setup}
    \end{subfigure}
    \vspace{-2mm}
    \caption{\small (a) Overview of the terrain course including slopes, stairs, and stepping stones; key geometric dimensions are annotated in white. (b) Real-world robot setup with front and rear depth cameras.}
    \label{fig:terrains_setup}
    \vspace{-2mm}
\end{figure}

\textbf{Real-World Terrain Settings.}
\label{sec:exp_setup}
As shown in \Cref{fig:terrains_setup}, we evaluate on a customized terrain course and a long curvy staircase. The course consists of (i) a $20^\circ$ slope, (ii) 1.2\,m-wide staircases with 20\,cm step height and step lengths of 22\,cm and 25\,cm, and (iii) stepping stones (25\,cm side length, 60\,cm gaps). With G1's 21\,cm foot, these leave little margin for footholds. The long curvy staircase is with 20\,cm height and 30\,cm step length, additionally requiring non-zero angular velocity.

\textbf{Perception Settings.} During distillation, we render depth at 10\,Hz with FOV $101^\circ\times69^\circ$, raw resolution $240\times135$ downsampled to $48\times27$. We support up to $N_{\text{cam}}=4$ cameras (front/rear/left/right) for omni-directional locomotion in simulation. 
To minimize hardware modification on Unitree G1 (mounting holes only at front and rear), we deploy $N_{\text{cam}}=2$ in a front--back configuration (\Cref{fig:terrains_setup}~(b)), which suffices for bidirectional locomotion. We use ZED~2i cameras at 15\,Hz with \textit{neural\_light} mode, downsampled to $48\times27$ before policy input. 
Realistic depth randomization (process delay, noise, dropout) is summarized in Appendix~\ref{appendix:dr}.

\subsection{Scalable Multi-Depth Rendering Performance}
\begin{table*}[t]
\centering
\begingroup
\scriptsize
\resizebox{\textwidth}{!}{%
\begin{tabular}{l c c cc cc cc}
\toprule
\multirow{2}{*}{\textbf{Method}} &
\multirow{2}{*}{\textbf{Dyn. Mesh}} &
\multirow{2}{*}{\textbf{Num. Envs}} &
\multicolumn{2}{c}{\textbf{$N_{\text{cam}}=1$}} &
\multicolumn{2}{c}{\textbf{$N_{\text{cam}}=2$}} &
\multicolumn{2}{c}{\textbf{$N_{\text{cam}}=4$}} \\
\cmidrule(lr){4-5} \cmidrule(lr){6-7} \cmidrule(lr){8-9}
& & 
& VRAM (GB)$\downarrow$ & Iter. (s)$\downarrow$
& VRAM (GB)$\downarrow$ & Iter. (s)$\downarrow$
& VRAM (GB)$\downarrow$ & Iter. (s)$\downarrow$ \\
\midrule
IsaacGym PhysX  & \checkmark & 1024 & 16.8 & 35.6\pmval{4.1} & 22.5 & 70.1\pmval{8.3} & 33.9 & 146.5\pmval{16.1} \\
IsaacSim RTX   & \checkmark & 1024 & 17.5 & 5.3\pmval{0.1} & 23.2 & 7.6\pmval{0.1} & 34.4 & 12.6\pmval{0.1} \\
IsaacSim Warp  & \texttimes & 1024 & \textbf{12.8} & 3.5\pmval{0.1} & 15.1 & 5.9\pmval{0.1} & 20.7 & 9.1\pmval{0.1} \\
IsaacLab Warp 2.3.2 & \checkmark & 1024 & 15.0 & 6.6\pmval{0.1} & 17.8 & 8.5\pmval{0.1} & 23.5 & 14.7\pmval{0.1} \\
\textbf{Ours (Alg.~\ref{alg:depth_raycast})}
& \checkmark & 1024
& 13.3 & \textbf{1.3}\pmval{0.0}
& \textbf{14.6} & \textbf{1.5}\pmval{0.1}
& \textbf{17.3} & \textbf{1.9}\pmval{0.1} \\
\bottomrule
\end{tabular}}
\endgroup
\vspace{-2mm}
\caption{\small Depth rendering scalability across different numbers of depth cameras with $N_{\text{env}}=1024$.}
\label{tab:depth_render}
\vspace{-1mm}
\end{table*}

\begin{table*}[t]
\centering
\begingroup
\resizebox{0.99\textwidth}{!}{%
\begin{tabular}{c c cccc}
\toprule
\textbf{Task} & \textbf{$N_{\text{cam}}$~(Config.)}
& \textbf{Slopes}
& \textbf{Stairs Up}
& \textbf{Stairs Down}
& \textbf{Stepping Stones} \\
\midrule

\multirow{4}{*}{\textbf{Bidirectional}}
& \cellcolor{verylightgray}{\textit{Expert Level}}
& \cellcolor{verylightgray}{\textit{6.0}}
& \cellcolor{verylightgray}{\textit{6.0}}
& \cellcolor{verylightgray}{\textit{6.0}}
& \cellcolor{verylightgray}{\textit{6.0}} \\
\cmidrule(lr){2-6}
& 1 (Down)
& \textbf{6.0}\pmval{0.0} & \textbf{6.0}\pmval{0.0} & \textbf{5.9}\pmval{0.1} & 5.1\pmval{0.1} \\
& 2 (F + B)
& \textbf{6.0}\pmval{0.0} & \textbf{6.0}\pmval{0.0} & \textbf{6.0}\pmval{0.0} & \textbf{6.0}\pmval{0.0} \\
& 4 (F + B + L + R)
& \textbf{6.0}\pmval{0.0} & \textbf{6.0}\pmval{0.0} & \textbf{6.0}\pmval{0.0} & \textbf{6.0}\pmval{0.0} \\
\midrule

\multirow{4}{*}{\textbf{Omnidirectional}}
& \cellcolor{verylightgray}{\textit{Expert Level}}
& \cellcolor{verylightgray}{\textit{6.0}}
& \cellcolor{verylightgray}{\textit{6.0}}
& \cellcolor{verylightgray}{\textit{6.0}}
& \cellcolor{verylightgray}{\textit{5.6}} \\
\cmidrule(lr){2-6}
& 1 (Down)
& \textbf{6.0}\pmval{0.0} & \textbf{6.0}\pmval{0.0} & \textbf{5.9}\pmval{0.1} & 3.0\pmval{0.2} \\
& 2 (F + B)
& \textbf{6.0}\pmval{0.0} & \textbf{6.0}\pmval{0.0} & \textbf{5.9}\pmval{0.1} & \textbf{4.5}\pmval{0.1} \\
& 4 (F + B + L + R)
& \textbf{6.0}\pmval{0.0} & \textbf{6.0}\pmval{0.1} & \textbf{6.0}\pmval{0.0} & \textbf{4.6}\pmval{0.0} \\
\bottomrule
\end{tabular}
}
\endgroup
\caption{\small
\textbf{Terrain levels~$\uparrow$} achieved under different $N_{\text{cam}}$ for bidirectional and omnidirectional locomotion.
}
\label{tab:terrain_levels}
\vspace{-2mm}
\end{table*}

\begin{wrapfigure}[6]{R}{0.4\columnwidth}
    \centering
    \vspace{-5mm}
    \includegraphics[width=\linewidth]{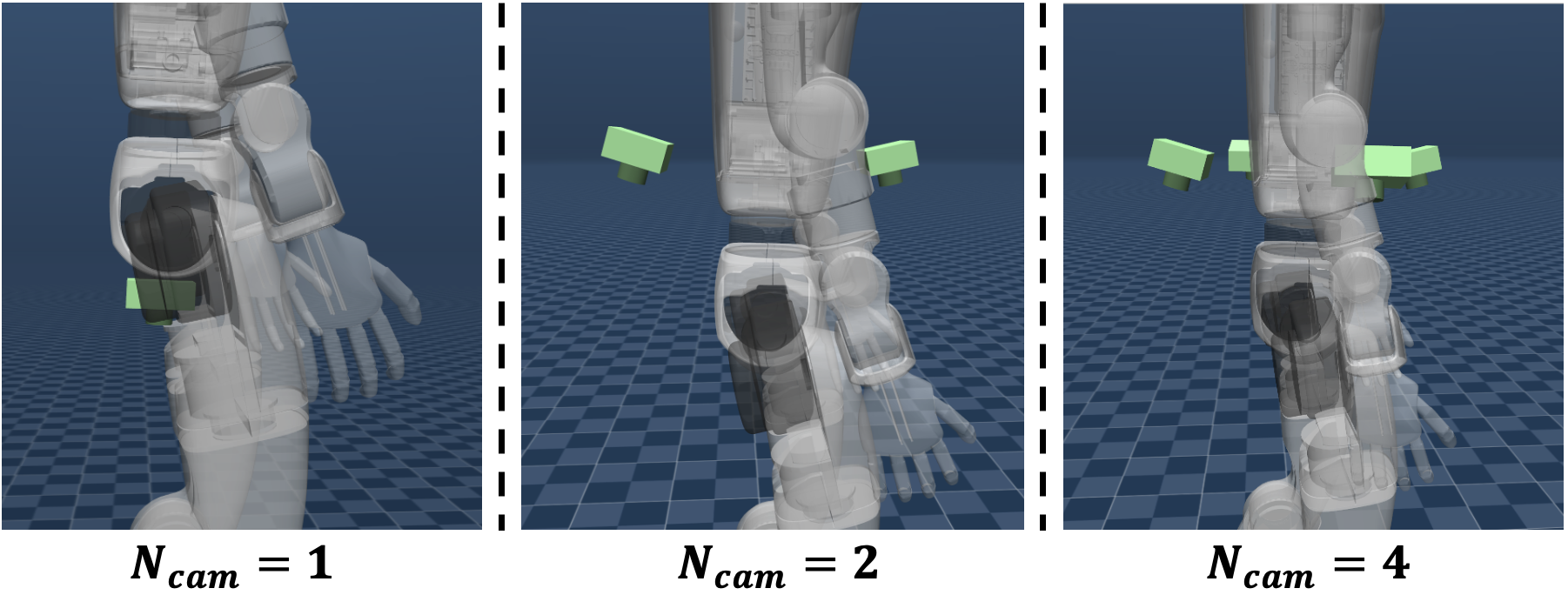}
    \vspace{-6mm}
    \caption{\small Camera poses for $N_{\text{cam}}=1,2,4$.}
    \label{fig:cameras}
\end{wrapfigure}

To answer \textbf{Q1}, we benchmark depth rendering pipelines for \method against IsaacGym PhysX, IsaacSim~\cite{mittal2023orbit}~RTX (TiledCamera), and IsaacSim Warp (RayCasterCamera).
As shown in~\Cref{tab:depth_render}, on a single L40S GPU with $N_{\text{cam}}=1,2,4$ over all terrains at $240\times135$ resolution, \method supports ray-casting against both dynamic and static meshes and achieves a $5\times$ speedup over the fastest baseline (IsaacSim Warp), which does not support dynamic meshes. IsaacLab 2.3.2's dynamic-mesh \texttt{MultiMeshRayCasterCamera} is $5.1/5.7/7.7\times$ slower at $1/2/4$ cameras (\Cref{tab:depth_render}). Iter. includes physics, rendering, and inference for all methods. Both use canonical-frame queries; \method's CUDA-graph replay and pre-allocated buffers avoid the baseline's per-camera launches, blocking synchronization, and per-step allocation.

To show the necessity of multiple cameras, we compare achieved terrain levels for $N_{\text{cam}}=1,2,4$ under bidirectional and omnidirectional locomotion (\Cref{tab:terrain_levels}); camera configurations are in~\Cref{fig:cameras} ($N_{\text{cam}}=1$ uses a downward-facing camera). Performance degrades sharply on stepping stones with fewer cameras, where 70\,cm gaps demand a wide FOV aligned with the walking direction. Reliable multi-directional locomotion thus requires at least two cameras covering each walking direction.

In single-shot per-terrain simulation tests, Hiking-in-the-Wild~\cite{zhu2026hiking} traverses slopes bidirectionally and stairs forward but fails on stepping stones; \method traverses all three. These outcomes are not aggregate success-rate estimates.

\subsection{Robustness to Asymmetric Depth Inputs, Unseen Terrain Widths and Payloads}
\begin{figure*}[t]
    \centering
    \includegraphics[width=0.98\textwidth]{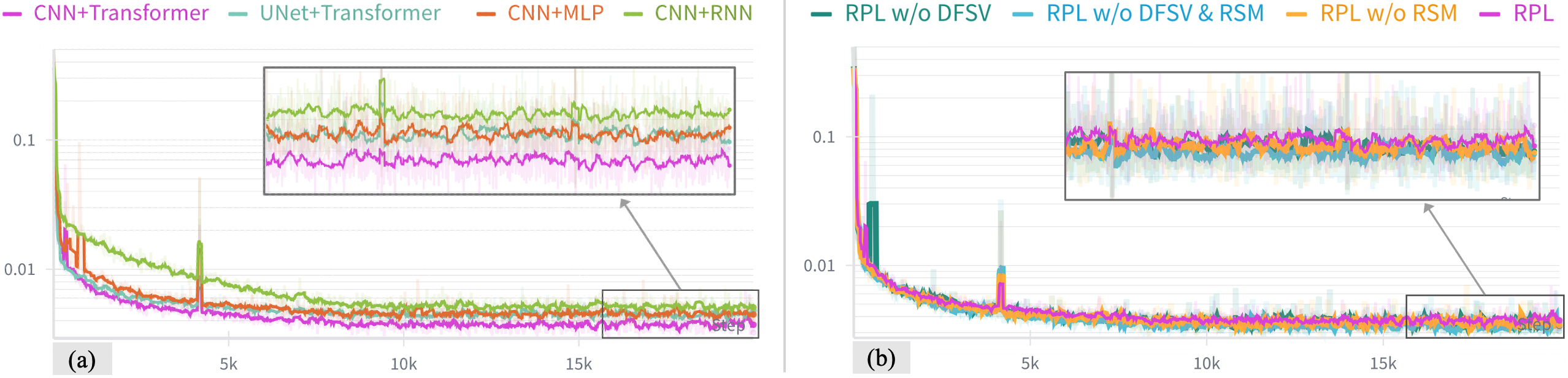}
    \vspace{-2mm}
    \caption{\small
    \textbf{Distillation Loss Comparison.} (a) compares among different network architectures. (b) compares among (1) \method; (2) \method w/o DFSV \& RSM; (3) \method w/o DFSV; and (4) \method w/o RSM.}
    \label{fig:distill}
    \vspace{-4mm}
\end{figure*}
\begin{figure}[t]
    \centering
    \begin{minipage}[c]{0.5\columnwidth}
        \centering
        \includegraphics[width=\linewidth]{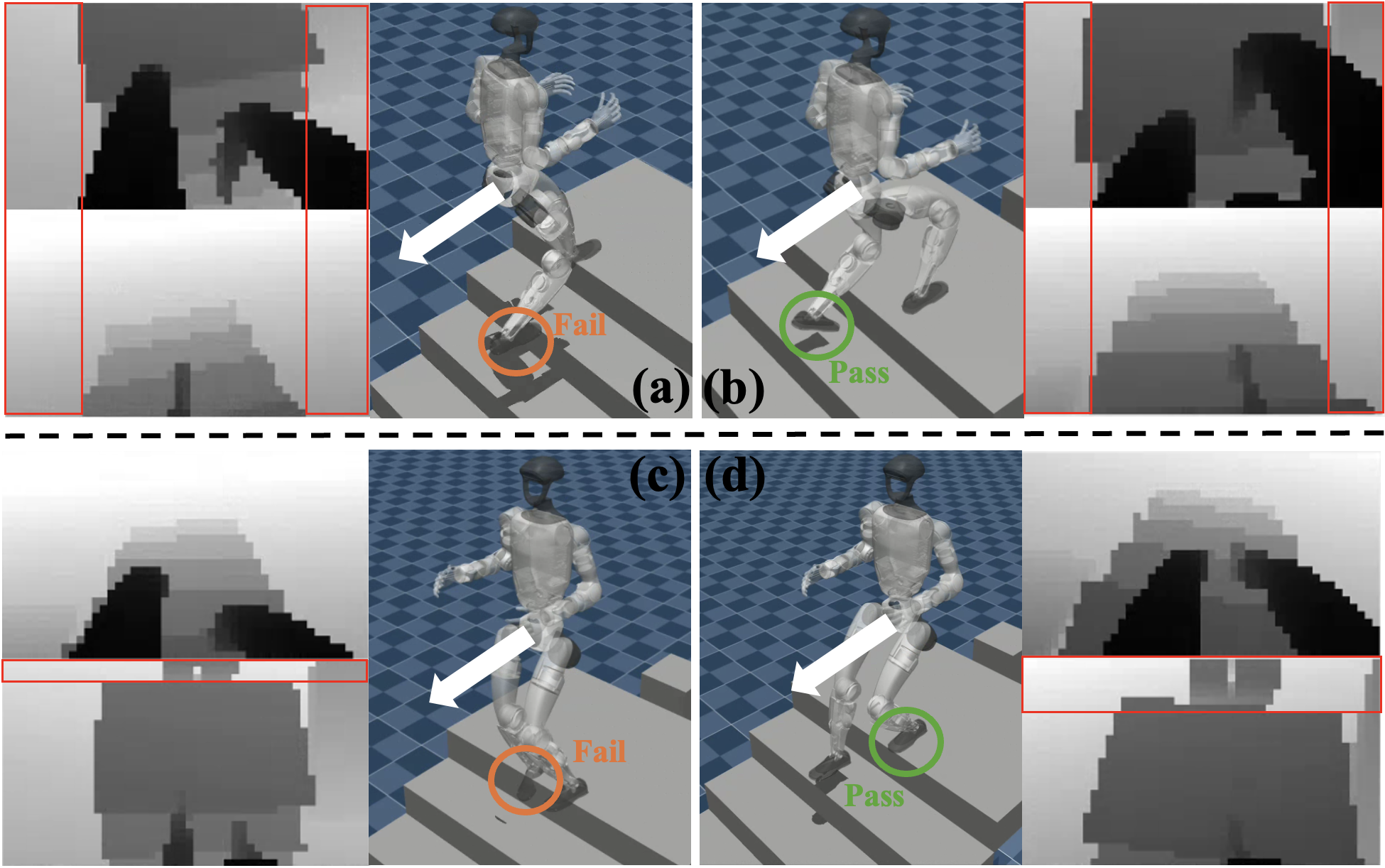}
        \captionof{figure}{\small \textbf{Bidirectional locomotion on unseen-width stairs with asymmetric depth observations}. (a) \method w/o RSM; (c) \method w/o DFSV; (b,d) \method.}
        \label{fig:mujoco_tricks}
    \end{minipage}\hfill
    \begin{minipage}[c]{0.46\columnwidth}
        \centering
        \small
        \setlength{\tabcolsep}{3pt}
        \renewcommand{\arraystretch}{1.05}
        \begin{tabular}{lcc}
        \toprule
        \textbf{Method} & \textbf{Forward} & \textbf{Backward} \\
        \midrule
        \method (CNN+Transformer)  & \goodnumber{10/10} & \goodnumber{10/10} \\
        \;\; w/o DFSV              & 0/10 & 2/10 \\
        \;\; w/o RSM               & 2/10 & 0/10 \\
        \;\; w/o DFSV \& RSM       & 0/10 & 0/10 \\
        \;\; w/o force pert.       & 3/10 & 3/10 \\
        \midrule
        CNN+MLP                    & 6/10 & 7/10 \\
        CNN+RNN                    & 5/10 & 5/10 \\
        U-Net+Transformer          & 1/10 & 0/10 \\
        \bottomrule
        \end{tabular}
        \captionof{table}{\small \textbf{Whole terrain course traversal success rate}.
        Top: DFSV/RSM ablations. Bottom: alternative network architectures, all with force perturbation, DFSV and RSM if applicable.}
        \label{tab:traversal_success}
    \end{minipage}
    \vspace{-6mm}
\end{figure}

To answer \textbf{Q2}, we ablate RSM, DFSV with four variants: \method, \method w/o RSM, \method w/o DFSV, and \method w/o DFSV~\&~RSM. All four reach similar final distillation losses (\Cref{fig:distill}~(b)), indicating the masked inputs still preserve enough task-relevant information for action supervision. Despite comparable training losses, deployment behavior differs substantially.
\begin{wrapfigure}[10]{R}{0.42\columnwidth}
    \centering
    \vspace{-4mm}
    \includegraphics[width=\linewidth]{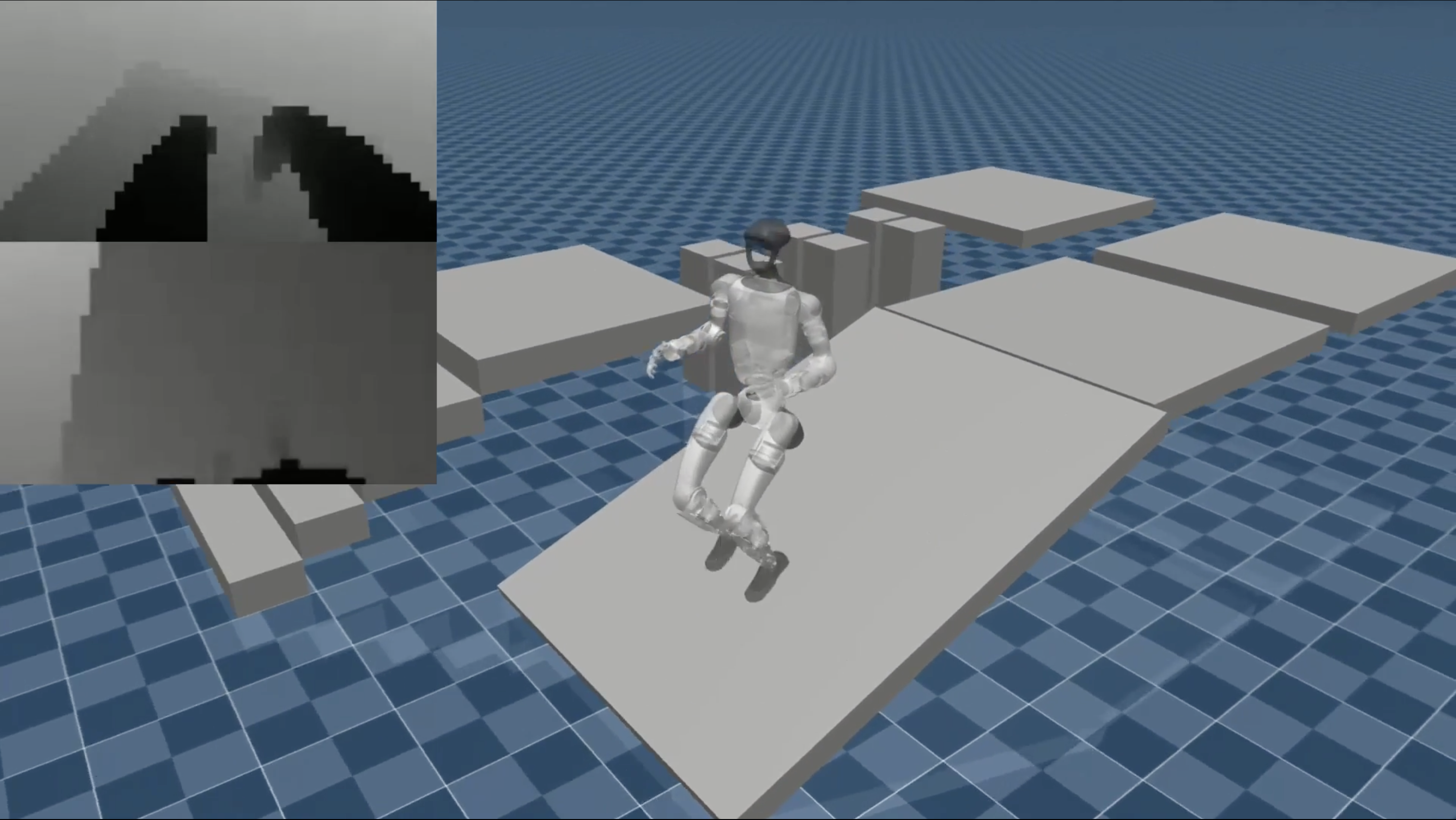}
    \caption{\small Terrain course in simulation.}
    \label{fig:mujoco_terrain}
    \vspace{-4mm}
\end{wrapfigure}

We visualize sim-to-sim rollouts in MuJoCo~\cite{mujoco} on a terrain course consisting of (i) a $25^\circ$ slope, (ii) 1.2\,m-wide staircases with 20\,cm step height and step lengths of 25\,cm, and (iii) stepping stones (25\,cm side length, 60\,cm gaps). The weight of the fake hand is changed to 2.5kg for the real world loco-manipulation setting in~\Cref{fig:teaser}~(b).
During backward descent (a,b), \method w/o RSM fails on the unseen narrow stairs while \method remains stable.
During forward descent (c,d), the front camera sees stairs while the rear still perceives stepping stones, creating asymmetric multi-view inputs; without DFSV, the policy is distracted by irrelevant rear-view cues and falls.
The whole-course traversal success rates in~\Cref{tab:traversal_success} (top block) confirm this: removing DFSV, RSM, or both drops success from 10/10 to as low as 0/10, and removing the Stage-1 force perturbation drops it to 3/10 under the 2.5\,kg payload. RSM, DFSV, and the force perturbation thus jointly enable robust bidirectional locomotion under OOD geometry, asymmetric perception, and end-effector payloads.

\subsection{Training Architecture and Distillation Ablations}
To answer \textbf{Q3}, we compare four distillation backbones (\Cref{fig:distill}~(a)): (i) \textit{CNN+MLP}, (ii) \textit{CNN+RNN}, (iii) \textit{CNN+Transformer} (\textbf{Ours}), and (iv) \textit{U-Net+Transformer} that reconstructs a height map via $\mathcal{L}_{\text{rec}} = \mathbb{E}_t[\|f_{\text{rec}}(\mathbf{D}_t) - \mathbf{H}_t\|_2^2]$ jointly with~\Cref{eq:distill_loss}. \textit{CNN+Transformer} achieves the lowest distillation loss \emph{and} the highest deployment success (\Cref{tab:traversal_success}, bottom block) thanks to its attention-based multi-view fusion. While~\Cref{sec:rsm} shows similar losses do not guarantee similar robustness, with DFSV and RSM applied, higher loss here reliably indicates worse deployment—so the loss remains a useful coarse selection signal. U-Net's pixel-wise reconstruction is also incompatible with RSM, limiting deployment robustness under partial observations.

\subsection{Multi-Terrain Gradient Analysis -- DAgger Only vs. Hybrid DAgger + RL}
\label{appendix:dagger_vs_hybrid}

\begin{figure}[h]
    \centering
    \includegraphics[width=\textwidth]{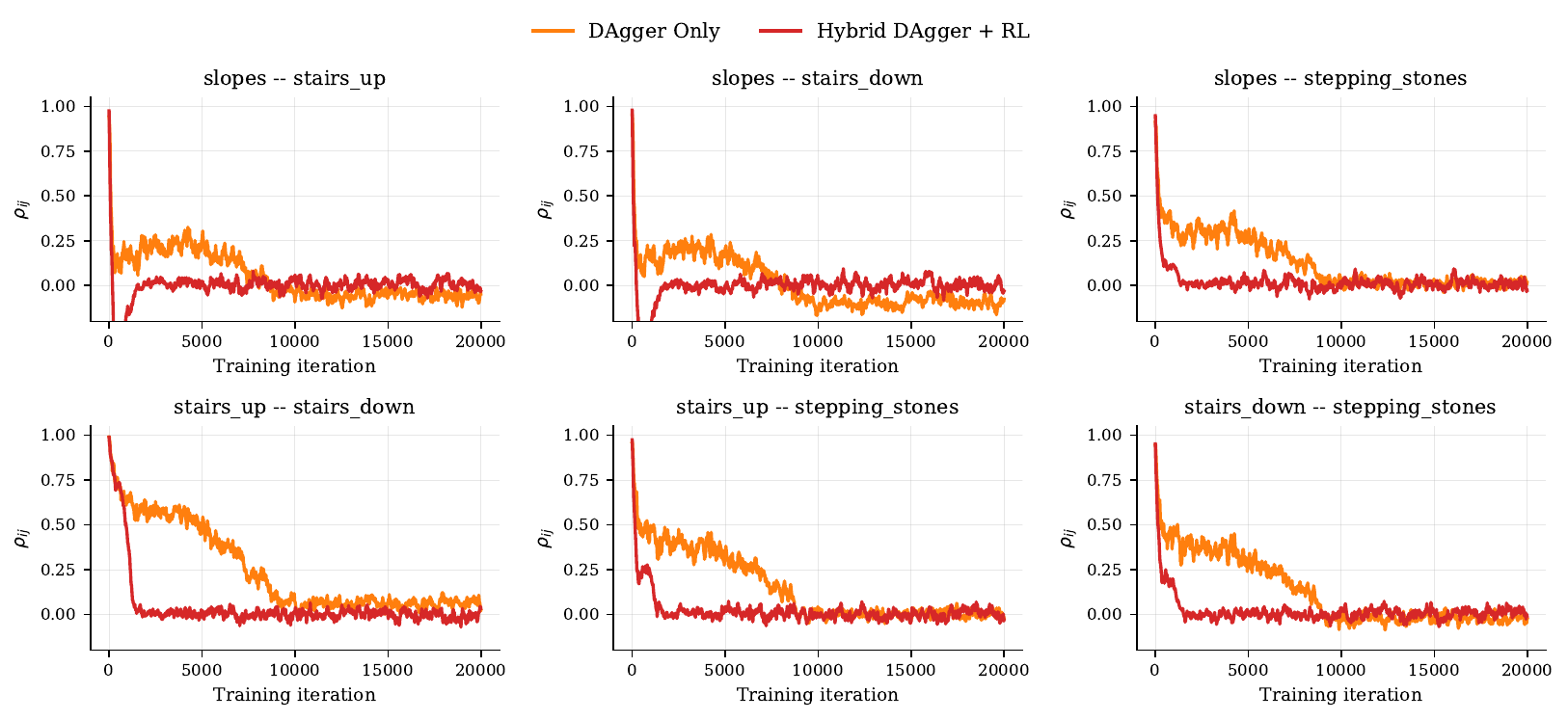}
    \caption{\small \textbf{Pairwise inter-terrain gradient cosine during training} for the $\binom{4}{2}=6$ terrain pairs. DAgger Only (orange) preserves higher cosine throughout, while Hybrid DAgger\,+\,RL (red) collapses toward $0$ within the first few hundred iterations.}
    \label{fig:gradient_cos}
\end{figure}

\begin{figure}[h]
    \centering
    \includegraphics[width=\textwidth]{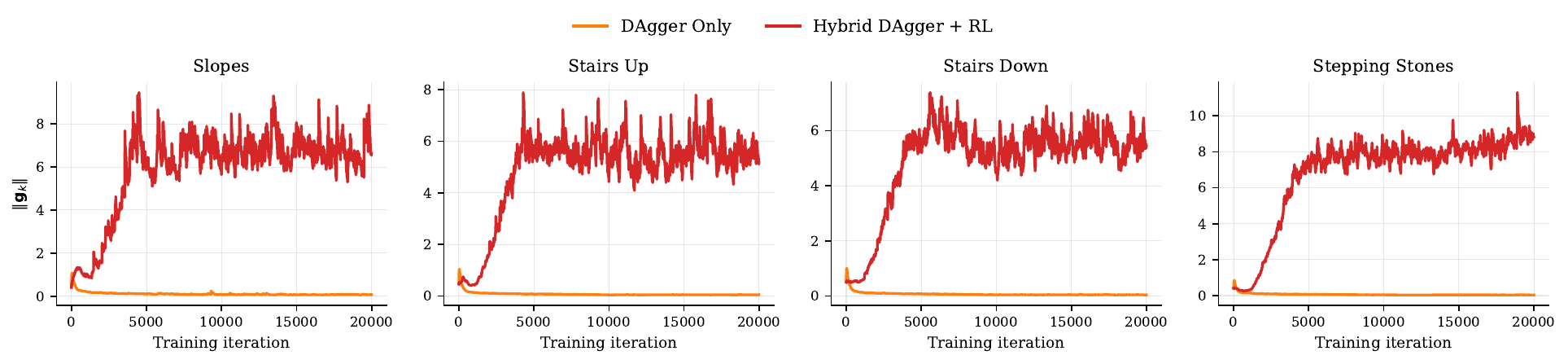}
    \caption{\small \textbf{Per-terrain gradient norm $\|\mathbf{g}_k\|$.} Hybrid's norms (red) remain $>0$ throughout training and are comparable to or larger than DAgger Only's (orange), so the cosine collapse in~\Cref{fig:gradient_cos} is real decorrelation, not vanishing gradients.}
    \label{fig:gradient_norm}
\end{figure}

\begin{figure}[h]
    \centering
    \includegraphics[width=\textwidth]{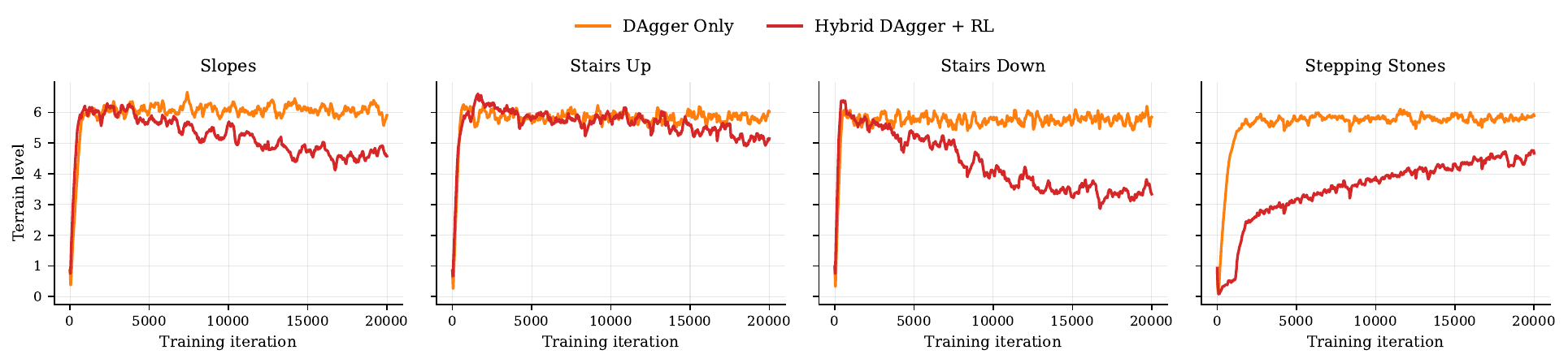}
    \caption{\small \textbf{Terrain-level curves under DAgger Only vs.\ Hybrid DAgger\,+\,RL.} DAgger Only (orange) reaches and sustains a higher terrain level on every terrain family, while Hybrid DAgger\,+\,RL (red) converges more slowly and plateaus at lower levels.}
    \label{fig:dagger_vs_hybrid_terrain}
\end{figure}

We compare two distillation objectives: (i) \emph{DAgger Only}, training the student only with the BC action regression loss in~\Cref{eq:distill_loss}; and (ii) \emph{Hybrid DAgger\,+\,RL} as proposed by~\cite{wu2026php}, which adds a PPO-style policy-gradient term so the student can both imitate the expert and refine itself with task rewards:
\begin{equation}
\mathcal{L}_{\text{hybrid}} = \lambda_{\text{PPO}}\,\mathcal{L}_{\text{PPO}} + \lambda_{\text{D}}\,\mathcal{L}_{\text{distill}},\quad \lambda_{\text{PPO}} + \lambda_{\text{D}} = 1,
\label{eq:hybrid_loss}
\end{equation}
where $\mathcal{L}_{\text{distill}}$ is the BC term from~\Cref{eq:distill_loss} and $\mathcal{L}_{\text{PPO}}$ is a standard PPO actor objective. We tune down $\lambda_{\text{D}}$ gradually following a cosine scheduling scheme, which is capped at 0.1. The hybrid is appealing in principle—the policy-gradient term could let the student improve in regions where the height-map expert is suboptimal under the new depth-only observation.

\textbf{Higher terrain levels.} On every terrain family, DAgger Only reaches a substantially higher terrain level and converges faster than Hybrid DAgger\,+\,RL (\Cref{fig:dagger_vs_hybrid_terrain}).

\textbf{Better gradient alignment.} Let $\mathcal{B}_k$ denote the subset of a shared mini-batch belonging to terrain $k\in\{1,\ldots,K\}$ ($K{=}4$ here). For each terrain we compute the per-terrain actor gradient
\begin{equation}
\mathbf{g}_k \;=\; \nabla_\theta\,\mathcal{L}_k(\mathcal{B}_k),
\qquad
\mathcal{L}_k =
\begin{cases}
\mathcal{L}_{\text{distill}}^{(k)} & \text{(DAgger Only)},\\[2pt]
\lambda_{\text{PPO}}\,\mathcal{L}_{\text{PPO}}^{(k)} + \lambda_{\text{D}}\,\mathcal{L}_{\text{distill}}^{(k)} & \text{(Hybrid DAgger\,+\,RL)},
\end{cases}
\label{eq:per_terrain_grad}
\end{equation}
and the pairwise cosine between any two terrains $i,j$:
\begin{equation}
\rho_{ij} \;=\; \frac{\langle \mathbf{g}_i,\,\mathbf{g}_j\rangle}{\|\mathbf{g}_i\|\,\|\mathbf{g}_j\|}.
\label{eq:pair_cos}
\end{equation}
A high $\rho_{ij}$ ($\rho_{ij}\!\to\!1$) means updating $\theta$ to reduce loss on terrain $i$ also reduces loss on terrain $j$, so the parameters move in a direction shared across terrains and one step improves multiple terrains at once; $\rho_{ij}\!\approx\!0$ means the two updates are roughly orthogonal, so each step pulls $\theta$ in its own direction and only improves one terrain at a time. We log $\rho_{ij}$ for all $\binom{K}{2}=6$ pairs (\Cref{fig:gradient_cos}). DAgger Only keeps $\rho_{ij}$ well above zero during early training, so each step is shared across terrains and the policy quickly learns the common locomotion fundamentals (balance, gait timing, depth--proprio fusion). Hybrid DAgger\,+\,RL drives $\rho_{ij}\!\to\!0$ within the first few thousand iterations (the per-terrain norms $\|\mathbf{g}_k\|$ stay non-zero in~\Cref{fig:gradient_norm}, so the gradients are just mutually orthogonal): the four terrains compete for capacity, per-step progress per terrain is smaller, and convergence is slower with lower final terrain levels (\Cref{fig:dagger_vs_hybrid_terrain}). This is not a flaw of hybrid itself: although we apply largely shared reward terms across the four terrains, the RL objective is value-driven and pushes the policy toward what is locally optimal \emph{on each terrain}, so the per-terrain gradients still pull $\theta$ in different directions—the multi-task RL gradient-interference phenomenon studied in prior work~\cite{yu2020gradient,chen2018gradnorm,liu2021conflict}. Resolving it would require either explicit gradient-surgery techniques or further reward co-design. Since DAgger Only already performs strongly, we adopt it in \method without investing in such treatment.

\subsection{Robust Real-World Bidirectional Locomotion}
To answer \textbf{Q4}, we deploy \method on the Unitree G1 29 DoF humanoid for back-and-forth traversal across a 50\,m terrain course and a curved building staircase (\Cref{sec:exp_setup}). The transformer policy runs at 50\,Hz on an onboard Jetson Orin NX 16GB; depth capture, policy inference, and the Apple Vision Pro teleoperation pipeline communicate via shared-memory buffers to reduce system latency. We run 10 whole-course bidirectional traversals (5 without payload, 5 with a 2\,kg payload), where a single misstep on any subterrain ends the trial.

\begin{wraptable}[8]{r}{0.55\columnwidth}
    \centering
    \vspace{-5mm}
    \footnotesize
    \setlength{\tabcolsep}{3pt}
    \renewcommand{\arraystretch}{0.95}
    \resizebox{\linewidth}{!}{\begin{tabular}{lcccc}
    \toprule
    \multirow{2}{*}{\textbf{Setting}} & \multirow{2}{*}{\textbf{Suc.}} & \multicolumn{3}{c}{\textbf{Failure}} \\
    \cmidrule(lr){3-5}
    & & (a) Stones & (b) Stairs & (c) Payload \\
    \midrule
    w/o payload & \goodnumber{3/5} & 1 & 1 & --- \\
    w/ 2\,kg payload & \goodnumber{3/5} & 1 & 0 & 1 \\
    \midrule
    \textbf{Total} & \textbf{6/10} & 2 & 1 & 1 \\
    \bottomrule
    \end{tabular}}
    \captionsetup{width=\linewidth}
    \caption{\footnotesize \textbf{Real-world whole-course traversal} (10 trials). Failure modes (a)-(c) detailed in~\Cref{appendix:real_world}.}
    \label{tab:real_world_success}
    \vspace{-3mm}
\end{wraptable}

As shown in the \textbf{supplementary video},~\Cref{fig:teaser}, and~\Cref{tab:real_world_success}, \method achieves an overall 6/10 whole-course success rate (3/5 without payload, 3/5 with the 2\,kg payload), each successful trial chaining a $20^\circ$ slope, staircases of two step lengths (22\,/\,25\,cm), and 25\,cm stepping stones with 60\,cm gaps in either forward or backward direction for $\sim$200\,m and $\sim$2\,min of continuous traversal. The robot also bends down to pick up the 2\,kg payload and transports it across the entire course, and generalizes zero-shot to an in-the-wild curved building staircase (30\,cm step) with unseen curvature and lighting. The remaining failures concentrate in three categories—foot-edge contacts on stepping stones and on the lateral side edge of the staircase, and payload swing—analyzed in detail in~\Cref{appendix:real_world}. 
Overall, these results demonstrate robustness under long horizons, asymmetric observations, narrow terrains, and payloads on a real humanoid.

\subsection{Real-World Study and Failure Mode Analysis}
\label{appendix:real_world}

\begin{figure}[h]
    \centering
    \includegraphics[width=\textwidth]{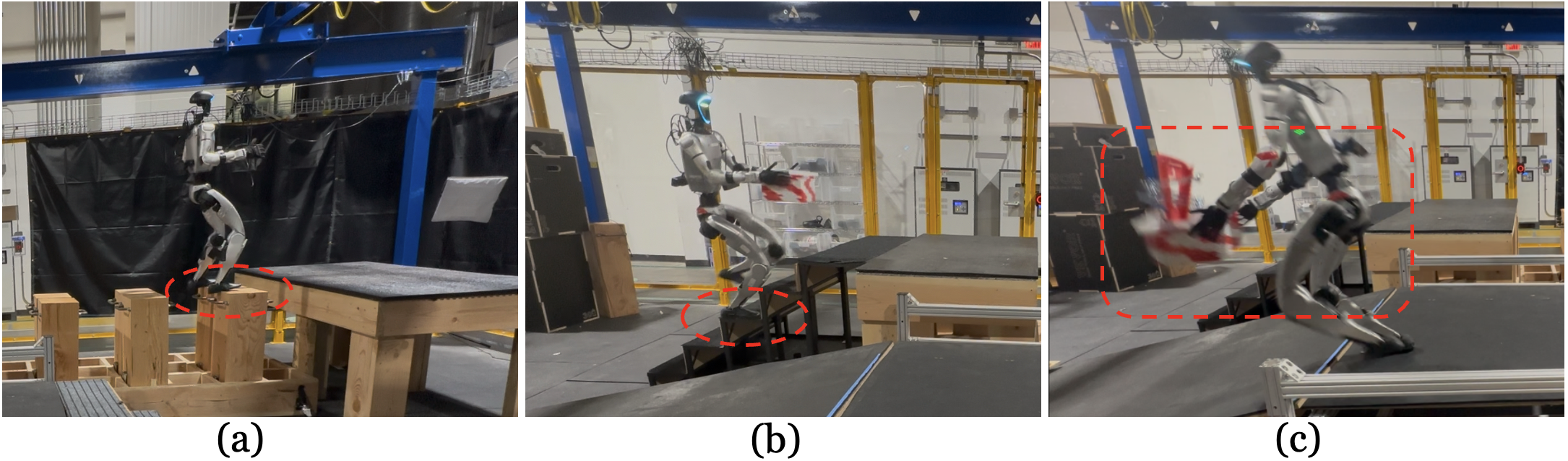}
    \caption{\small \textbf{Real-world failure modes of \method.} Representative failure cases observed during real-world deployment: (a) foot landings near the perimeter of stepping stones; (b) foot landings near the lateral side edge of the staircase; (c) payload swing from the imperfect hook-style end-effector attachment.}
    \label{fig:real_world_failures}
    \vspace{-2mm}
\end{figure}

We further break down the failure modes of the 10 whole-course trials reported in~\Cref{tab:real_world_success} (6/10 overall). The remaining trials split across three categories (\Cref{fig:real_world_failures}):

\textbf{(a,b) Foot landings near terrain edges.} The 21\,cm foot leaves little margin against the 25\,cm stones (a) and the 1.2\,m-wide staircase (b); small noise or drift pushes landings to the boundary. The Stage-1 foot-edge penalty mitigates but does not eliminate this; a stricter edge-avoidance reward would help.

\textbf{(c) Payload swing.} Our passive hook lets the payload swing, shifting the CoM beyond the Stage-1 force curriculum—the single payload-trial failure falls under this mode. Mitigations: a pinching gripper or an end-effector stabilization term~\cite{li2025hold}.

\section{Conclusion}
\label{sec:conclusion}
We propose \method, a two-stage learning framework for multi-directional humanoid perceptive locomotion over complex terrains and remain robust with payloads.
\method first trains terrain-specialized expert policies with privileged height-map observations to learn decoupled locomotion and manipulation, and then distills them into a single depth-based transformer policy that fuses multi-view depth inputs.
Our parallel depth renderer efficiently ray-casts dynamic robot and static terrain meshes with realistic latency, noise, and dropout, achieving $5\times$ speedup over existing simulation renderers.
We further introduce DFSV and RSM to improve robustness under asymmetric multi-view observations and unseen terrain widths. Extensive simulation and real-world experiments on the Unitree G1 demonstrate robust and long-horizon back-and-forth locomotion across challenging terrains.

\section{Limitations}
\label{sec:limitations}
Despite its strong performance, \method has two main limitations. First, we do not demonstrate real-world sideways locomotion in this paper, as achieving high-level sideways performance on discrete terrains such as stepping stones, remains non-trivial for both the expert training and the distillation.
Second, while DFSV improves robustness with fixed cameras but does not explicitly learn active viewpoint selection for highly occluded or ambiguous loco-manipulation scenarios.



\clearpage

\bibliography{references}

\clearpage

\appendix
\section{Appendix}
\subsection{Notation Summary}
\label{appendix:notation}
\begin{table}[H]
\caption{Notation summary.}
\label{tab:symbols}
\centering
\small
\setlength{\tabcolsep}{10pt}
\renewcommand{\arraystretch}{1.15}
\begin{tabular}{ c c }
\hline
Symbol & Meaning \\
\hline

\multicolumn{2}{c}{\textbf{Proprioception}} \\
\hline
$\boldsymbol{s}_t^p$ & Proprioceptive history \\
$\boldsymbol{q}_{t-4:t}$ & Joint positions history \\
$\dot{\boldsymbol{q}}_{t-4:t}$ & Joint velocities history \\
$\boldsymbol{\omega}^{\text{root}}_{t-4:t}$ & Root angular velocity history \\
$\boldsymbol{g}_{t-4:t}$ & Gravity vector history \\
$\boldsymbol{a}_{t-5:t-1}$ & Action history \\
\hline

\multicolumn{2}{c}{\textbf{Goals}} \\
\hline
$\boldsymbol{G}_t^l$ & Lower-body locomotion goals \\
$\boldsymbol{v}_t^{\text{lin}}$ & Linear velocity command \\
$\boldsymbol{w}_t^{\text{yaw}}$ & Angular velocity command \\
$\boldsymbol{\phi}_t^{\text{stance}}$ & Stance/walking mode command \\
$\boldsymbol{h}_t^{\text{root}}$ & Root height command \\
$\boldsymbol{o}_t^{\text{torso}}$ & Torso orientation command \\
$\boldsymbol{G}_t^u$ & Upper-body manipulation goals \\
$\boldsymbol{q}_{\text{upper},t}$ & Upper-body joint target \\
\hline

\multicolumn{2}{c}{\textbf{Perceptual Observation}} \\
\hline
$\boldsymbol{H}_t$ & Privileged height map \\
$\boldsymbol{D}_t$ & Multi-view depth observation \\
$\boldsymbol{d}_t^{(i)}$ & Depth image from camera $i$ at time $t$ \\
$N_{\text{cam}}$ & Number of depth cameras \\
\hline
\end{tabular}
\end{table}

\subsection{Domain Randomization}
\label{appendix:dr}
\begin{table}[H]
\caption{Domain randomization terms during distillation.}
\label{tab:dr}
\centering
\begin{tabular}{ c c }
\hline
Term           & Value                              \\ 
\hline
\multicolumn{2}{c}{\textbf{Dynamics Randomization}}  \\ 
\hline
Friction & $\mathcal{U}(0.5, 1.25)$            \\
Link mass    & $\mathcal{U}(0.9, 1.2) \times \text{default} \ \text{kg}$            \\
Base mass    & $\mathcal{U}(-1.0, 3.0) \ \text{kg}$            \\
Base COM range & $x \sim \mathcal{U}(-0.025, 0.025)$\text{m} \\
               & $y \sim \mathcal{U}(-0.05, 0.05)$\text{m} \\
               & $z \sim \mathcal{U}(-0.05, 0.05)$\text{m} \\
Control delay      & $\mathcal{U}(0, 20)\text{ms}$           \\ 

\hline
\multicolumn{2}{c}{\textbf{Perception Randomization (Depth Cameras)}} \\
\hline
Depth cam translation &
$x \sim \mathcal{U}(-0.025, 0.025)\ \text{m}$ \\
& $y \sim \mathcal{U}(-0.025, 0.025)\ \text{m}$ \\
& $z \sim \mathcal{U}(-0.025, 0.025)\ \text{m}$ \\
Depth cam rotation (Euler) &
$\phi \sim \mathcal{U}(-2.5, 2.5)^\circ$ \\
& $\theta \sim \mathcal{U}(-3.0, 3.0)^\circ$ \\
& $\psi \sim \mathcal{U}(-2.5, 2.5)^\circ$ \\
Depth cam FOV &
$\Delta \mathrm{FOV} \sim \mathcal{U}(-2.0, 2.0)^\circ$ \\
Pixel dropout & $p_{\mathrm{drop}} = 0.05$ \\
Depth noise & $\sigma_d = 0.1 \cdot \text{depth}$ \\
\hline
\end{tabular}
\end{table}

\end{document}